\documentclass[10pt,twocolumn,letterpaper]{article}

\usepackage{cvpr}
\usepackage{times}
\usepackage{graphicx}
\usepackage{amsmath}
\usepackage{amssymb}
\usepackage{amsthm}
\usepackage{algorithm}
\usepackage{algorithmic}
\usepackage{wasysym}
\usepackage{macro}

\usepackage[pagebackref=true,breaklinks=true,letterpaper=true,colorlinks,bookmarks=false]{hyperref}

\cvprfinalcopy

\ifcvprfinal\fi
\begin{document}

\title{Deep Hashing: A Joint Approach for Image Signature Learning}

\author{Yadong Mu\\
Institute of Computer Science and Technology\\
Peking University, China\\
{\tt\small myd@pku.edu.cn}
\and
Zhu Liu\\
Multimedia Department\\
AT\&T Labs, U.S.A.\\
{\tt\small zliu@researh.att.com}
}

\maketitle

\begin{abstract}
Similarity-based image hashing represents crucial technique for visual data storage reduction and expedited image search. Conventional hashing schemes typically feed hand-crafted features into hash functions, which separates the procedures of feature extraction and hash function learning. In this paper, we propose a novel algorithm that concurrently performs feature engineering and non-linear supervised hashing function learning. Our technical contributions in this paper are two-folds: 1) deep network optimization is often achieved by gradient propagation, which critically requires a smooth objective function. The discrete nature of hash codes makes them not amenable for gradient-based optimization. To address this issue, we propose an exponentiated hashing loss function and its bilinear smooth approximation. Effective gradient calculation and propagation are thereby enabled; 2) pre-training is an important trick in supervised deep learning. The impact of pre-training on the hash code quality has never been discussed in current deep hashing literature. We propose a pre-training scheme inspired by recent advance in deep network based image classification, and experimentally demonstrate its effectiveness. Comprehensive quantitative evaluations are conducted on several widely-used image benchmarks. On all benchmarks, our proposed deep hashing algorithm outperforms all state-of-the-art competitors by significant margins. In particular, our algorithm achieves a near-perfect 0.99 in terms of Hamming ranking accuracy with only 12 bits on MNIST, and a new record of 0.74 on the CIFAR10 dataset. In comparison, the best accuracies obtained on CIFAR10 by existing hashing algorithms without or with deep networks are known to be 0.36 and 0.58 respectively.
\end{abstract}

\section{Introduction}

Recent years have witnessed spectacular progress on similarity-based hash code learning in a variety of computer vision tasks, such as image search~\cite{Chum08a}, object recognition~\cite{TorralbaFW08} and local descriptor compression~\cite{Strecha12} etc. The hash codes are highly compact (\eg, several bytes for each image) in most cases, which significantly reduces the overhead of storing visual big data and also expedites similarity-based image search. The theoretic ground of similarity-oriented hashing is rooted from Johnson-Lindenstrause theorem~\cite{dasguptaG03}, which elucidates that for arbitrary $n$ samples, some $\mathcal{O}(\log(n))$-dimensional subspace exists and can be found in polynomial time complexity. When embedded into this subspace, pairwise affinities among these $n$ samples are preserved with tight approximation error bounds. This seminal theoretic discovery sheds light on trading similarity preservation for high compression of large data set. The classic locality-sensitive hashing (LSH)~\cite{Indyk98} is a good demonstration for above tradeoff, instantiated in various similarity metrics such as Hamming distance~\cite{Indyk98}, cosine similarity~\cite{charikar02}, $\ell_p$ distance with $p \in (0,2]$~\cite{datar04}, Jaccard index~\cite{Broder98} and Euclidean distance~\cite{AndoniINR13}.

Images are often accompanied with supervised information in various forms, such as semantically similar~/~dissimilar data pairs. Supervised hash code learning~\cite{mu10,WangKC12} harnesses such supervisory information during parameter optimization and has demonstrated superior image search accuracy compared with unsupervised hashing algorithms~\cite{andoniI08,WeissTF08,GongLGP13}. Exemplar supervised hashing schemes include LDAHash~\cite{Strecha12}, two-step hashing~\cite{LinSSHS14}, and kernel-based supervised hashing~\cite{liuw12} etc.

Importantly, two factors are known to be crucial for hashing-based image search accuracy: the discriminative power of the features and the choice of hashing functions. In a typical pipeline of existing hashing methods, these two factors are separately treated. Each image is often represented by a vector of hand-crafted visual features (such as SIFT-based bag-of-words feature or sparse codes). Regarding hash functions, a large body of existing works have adopted linear functions owing to the simplicity. More recently, researchers have also explored a number of non-linear hashing functions, such as anchor-based kernalized hashing function~\cite{liuw12} and decision tree based function~\cite{LinSSHS14}.

This paper attacks the problem of supervised hashing by concurrently conducting visual feature engineering and hash function learning. Most of existing image features are designated for general computer vision tasks. Intuitively, by unifying these two sub-tasks in the same formulation, one can expect the extracted image features to be more amenable for the hashing purpose. Our work is inspired by recent prevalence and success of deep learning techniques~\cite{Lecun98gradient-basedlearning,Bengio09,KrizhevskySH12}. Though the unreasonable effectiveness of deep learning has been successfully demonstrated in tasks like image classification~\cite{KrizhevskySH12} and face analysis~\cite{SunWT14}, deep learning for supervised hashing still remains inadequately explored in the literature.

Salakhutdinov et al. proposed \emph{semantic hashing} in~\cite{SalakhutdinovH09}, where stacked Restricted Boltzmann Machines (RBMs) are employed for hash code generation. Nonetheless, the algorithm is primarily devised for indexing textual data and its extension to visual data is unclear. Xia et al.~\cite{XiaPLLY14} adopted a two-step hashing strategy similar to~\cite{LinSSHS14}. It firstly factorizes the data similarity matrix to obtain the target binary code for each image. In the next stage, the target codes and the image labels are jointly utilized to guide the network parameter optimization. Since the target codes are not updated once approximately learned in the first stage, the final model is only sub-optimal. Lai et al.~\cite{LaiPLY15} developed a convolutional deep network for hashing, comprised of shared sub-networks and a divide-and-encode module. However, the parameters of these two components are still separately learned. After the shared sub-networks are initialized, their parameters (including all convolutional/pooling layers) are frozen during optimizing the divide-and-encode module. Intrinsically, the method in~\cite{LaiPLY15} shall be categorized to two-step hashing, rather than simultaneous feature / hashing learning. Liong et al.~\cite{LiongLWMZ15} presented a binary encoding network built with purely fully-connected layers. The method essentially assumes that the visual features (\eg, GIST as used in the experiments therein) have been learned elsewhere and fed into its first layer as the input.

As revealed by above literature overview, a deep hashing method which simultaneously learns the features and hash codes remains missing in this research field, which inspires our work. The key contributions of this work include:
\begin{itemize}
\item We propose the first deep hashing algorithm of its kind, which performs concurrent feature and hash function learning over a unified network.

\item We investigate the key pitfalls in designing such deep networks. Particularly, there are two major obstacles: the gradient calculation from non-differentiable binary hash codes, and network pre-training in order to eventually stay at a ``good" local optimum. To address the first issue, we propose an exponentiated hashing loss function and devise its bilinear smooth approximation. Effective gradient calculation and propagation are thereby enabled. Moreover, an efficient pre-training scheme is also proposed. We verify its effectiveness through comprehensive evaluations on real-world visual data.

\item The proposed deep hashing method establishes new performance records on four image benchmarks which are widely used in this research area. For instance, on the CIFAR10 dataset, our method achieves a mean average precision of 0.73 for Hamming ranking based image search, which represents some drastic improvement compared with the state-of-the-art methods (0.58 for~\cite{LaiPLY15} and 0.36 for~~\cite{liuw12}).
\end{itemize}

\section{The Proposed Method}

\begin{figure*}[t]
\begin{center}
   \includegraphics[width=0.95\linewidth]{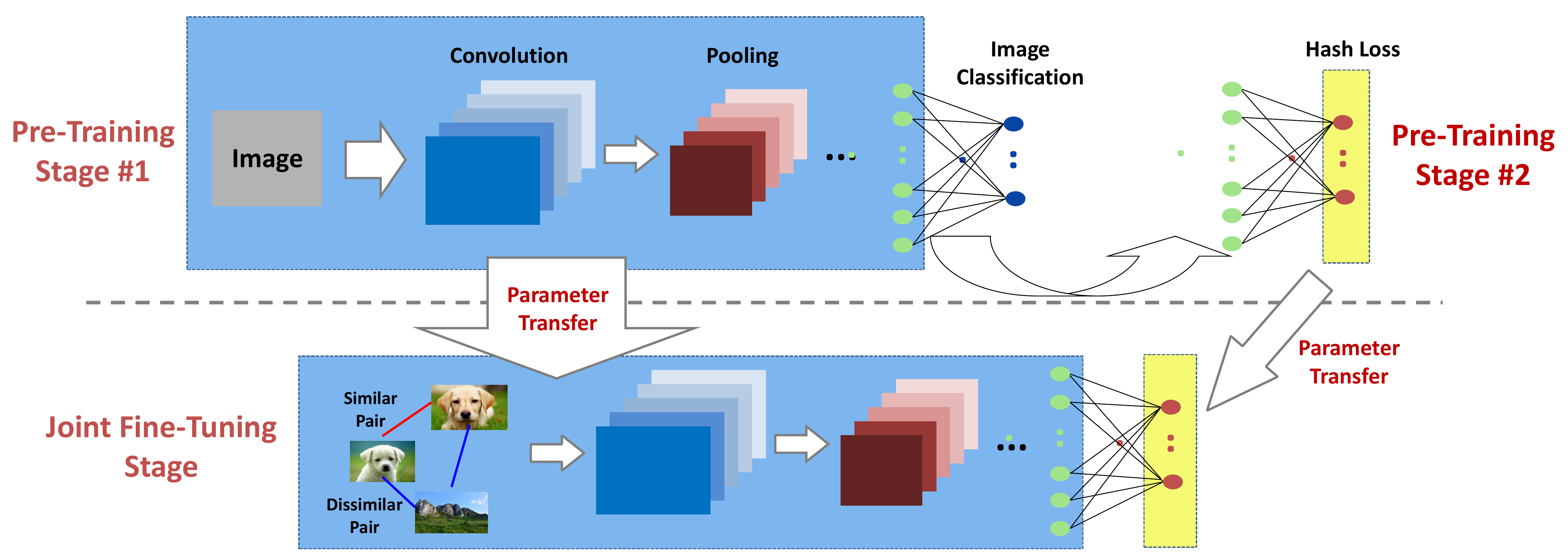}
\end{center}
   \caption{\small Illustration of our proposed deep network and the pre-training / fine-tuning process. Due to space limit, non-linear activation layers are not plotted in the diagram. See text for more explanations.}
\vspace{-0.15in}
\label{fig:network}
\end{figure*}

Throughout this paper we will use bold symbols to denote vectors or matrices, and italic ones for scalars unless otherwise instructed. Suppose a data set $\mathcal{X} = \{\x_1,\ldots,\x_n\}$ with supervision information is provided as the input. Prior works on supervised hashing have considered various forms of supervision, including triplet of items $\langle \x,\x^+,\x^- \rangle$ where the pair $\langle \x, \x^+ \rangle$ is more alike than the pair $\langle \x, \x^- \rangle$~\cite{mu10, norouzi12,LaiPLY15}, pairwise similar/dissimilar relations~\cite{liuw12} or specifying the label of each sample. Observing that triplet-type supervision incurs tremendous complexity during hashing function learning and semantic-level sample labels can be effortlessly converted into pairwise relations, hereafter the discussion focuses on supervision in pairwise fashion. Let $\mathcal{S}$, $\mathcal{D}$ collect all similar / dissimilar pairs respectively. For notational convenience, we further introduce a supervision matrix $\Y \in \{-1,0,1\}^{n \times n}$ as
\begin{eqnarray}
\Y_{i,j} = \left\{ \begin{array}{rl}
                  1, & (\x_i,\x_j) \in \mathcal{S} \\
                  -1, & (\x_i,\x_j) \in \mathcal{D} \\
                  0, & \mbox{otherwise}.
                \end{array}
\right. \label{eqn:y}
\end{eqnarray}

Figure~\ref{fig:network} illustrates our proposed pipeline of learning a deep convolutional network for supervised hashing. The network is comprised of two components: a topmost layer meticulously-customized for the hashing task and other conventional layers. The network takes a $p \times q$-sized images with $c$ channels as the inputs. The $K$ neurons on the top layer output either -1 or 1 as the hash code. Formally, each top neuron represents a hashing function $h_k(\x) : \mathbb{R}^{p \times q \times c} \mapsto \{-1, 1\},~ k=1\ldots K$, where $\x$ denotes the 3-D raw image. For notational clarity, let us denote the response vector on the second topmost layer as $\z = \phi(\x)$, where $\phi(\cdot)$ implicitly defines the highly non-linear mapping from the raw data to a specified intermediate layer.

For the topmost layer, we adopt a simple linear transformation, followed by a signum operation, which is formally presented as
\begin{eqnarray}
\vspace{-0.2in}
h_k(\x) = \mbox{sign}\left[ \w_k^\top \z \right] = \mbox{sign} \left[ \w_k^\top \phi(\x) \right].
\label{eqn:h}
\end{eqnarray}

The reminder of this section firstly introduces the hashing loss function and the calculation of its smoothed surrogate in Section~\ref{subsec:loss}. More algorithmic details of our proposed pretraining-finetuning procedure are delivered in Section~\ref{subsection:pretrain}.

\subsection{Exponentiated Code Product Optimization}
\label{subsec:loss}

The key purpose of supervised hashing is to elevate the image search accuracy. The goal can be intuitively achieved by generating discriminative hash codes, such that similar data pairs can be perfectly distinguished from dissimilar pairs according to the Hamming
distances calculated over the hash codes. A number of hashing loss functions have been devised by using above design principal. In particular, Norouzi et al.~\cite{NorouziF11} propose a hinge-like loss function. Critically, hinge loss is known to be non-smooth and thus complicates gradient-based optimization. Two other works~\cite{liuw12,LinSSHS14} adopt smooth $L_2$ loss defined on the inner product between hash codes.

It largely remains unclear for designing optimal hashing loss functions in perceptron-like learning. The major complication stems from the discrete nature of hash codes, which prohibits direct gradient computation and propagation as in typical deep networks. As such, prior works have investigated several tricks to mitigate this issue. Examples include optimizing a variational upper bound of the original non-smooth loss~\cite{NorouziF11}, or simply computing some heuristic-oriented sub-gradients~\cite{LaiPLY15}. In this work we advocate an exponential discrete loss function which directly optimizes the hash code product and enjoys a bilinear smoothed approximation. Compared with other alternative hashing losses, here we first show the proposed exponential loss arguably more amenable for mini-batch based iterative update and later exhibit its empirical superiority in the experiments.

Let $\mathbf{b}_i = \langle h_1(\x_i),\ldots,h_K(\x_i) \rangle^\top \in \{-1,1\}^K$ denote $K$ hash bits in vector format for data object $\x_i$. We also use the notations $\mathbf{b}_i(k)$, $\mathbf{b}_i(\setminus k)$ to stand for bit $k$ of $\mathbf{b}_i$ and the hash code with bit $k$ absent respectively. As a widely-known fact in the hashing literature~\cite{liuw12}, \emph{code product} admits a one-to-one correspondence to Hamming distance and comparably easier to manipulate. A normalized version of code product ranging over $[-1,1]$ is described as
\begin{eqnarray}
\mathbf{b}_i \circ \mathbf{b}_j = \textstyle \frac{1}{K} \sum_{k=1}^K \mathbf{b}_i(k) \mathbf{b}_j(k), \label{eqn:cp}
\vspace{-0.05in}
\end{eqnarray}
and when bit $k$ is absent, the code product using partial hash codes is
\begin{eqnarray}
\mathbf{b}_i(\setminus k) \circ \mathbf{b}_j(\setminus k) = \textstyle \mathbf{b}_i \circ \mathbf{b}_j - \frac{1}{K} \mathbf{b}_i(k)  \mathbf{b}_j(k). \label{eqn:partialcodeproduct}
\vspace{-0.05in}
\end{eqnarray}

\vspace{0.05in}
\noindent \textbf{Exponential Loss}: Given the observation that $\mathbf{b}_i \circ \mathbf{b}_j$ faithfully indicates the pairwise similarity, we propose to minimize an exponentiated objective function $\mathcal{Q}$ defined as the accumulation over all data pairs:
\begin{eqnarray}
\textstyle (\boldsymbol\theta^\ast,\w_k^\ast) = \arg \min_{\boldsymbol\theta,\w_k} \mathcal{Q} \triangleq \sum_{i,j} \ell(\x_i,\x_j),
\label{eqn:obj}
\end{eqnarray}
where $\boldsymbol\theta$ represents the collection of parameters in the deep networks excluding the hashing loss layer. The atomic loss term is
\begin{eqnarray}
\ell(\x_i,\x_j) = e^{ - \Y_{i,j} \left( \mathbf{b}_i \circ \mathbf{b}_j \right) }.
\label{eqn:aloss}
\vspace{-0.1in}
\end{eqnarray}

This novel loss function enjoys some elegant traits desired by deep hashing compared with those in BRE~\cite{BRE}, MLH~\cite{NorouziF11} and KSH~\cite{liuw12}. It establishes more direct connection to the hashing function parameters by maximizing the correlation of code product and pairwise labeling. In comparison, BRE and MLH optimize the parameters by aligning Hamming distance with original metric distances or enforcing the Hamming distance larger/smaller than pre-specified thresholds. Both formulations incur complicated optimization procedures, and their optimality conditions are unclear. KSH adopts a least-squares formulation for regressing code product onto the target labels, where a smooth surrogate for gradient computation is proposed. However, the surrogate heavily deviates from the original loss function due to its high non-linearity.

\vspace{0.1in}
\noindent \textbf{Gradient Computation}: A prominent advantage of exponential loss is its easy conversion into multiplicative form, which elegantly simplifies the derivation of its gradient. For presentation clarity, we hereafter only focus on the calculation conducted over the topmost hashing loss layer. Namely, $h_k(\x) = \mbox{sign}\left[ \w_k^\top \z \right]$ for bit $k$, where $\z=\phi(\x)$ are the response values at the second top layer and $\w_k$ are parameters to be learned for bit $k$ ($k=1,\ldots,K$).


Following the common practice in deep learning, two groups of quantities $\partial \mathcal{Q} / \partial \w_k$, $k=1\cdots K$ and $\partial \mathcal{Q} / \partial \z_i$ ($i$ ranges over the index set of current mini-batch) need to be estimated on the hashing loss layer at each iteration. The former group of quantities are used for updating $\w_k$, $k=1\cdots K$, and the latter are propagated backwards to the bottom layers. The additive algebra of hash code product in Eqn.~(\ref{eqn:cp}) inspires us
to estimate the gradients in a leave-one-out mode. For atomic loss in Eqn.~(\ref{eqn:aloss}), it is easily verified
\begin{eqnarray}
\ell(\x_i,\x_j) &=& e^{  - \Y_{i,j} \left( \mathbf{b}_i \circ \mathbf{b}_j \right)  }\nonumber \\
&=& \textstyle e^{ - \Y_{i,j} \left( \mathbf{b}_i(\setminus k) \circ \mathbf{b}_j(\setminus k) \right) } \cdot e^{ - \frac{1}{K} \Y_{i,j} \left( \mathbf{b}_i(k)  \mathbf{b}_j(k) \right) }, \nonumber
\end{eqnarray}
where only the latter factor is related to $\w_k$. Since the product $\mathbf{b}_i(k)  \mathbf{b}_j(k)$ can only be -1 or 1, we can linearize the latter factor through exhaustively enumerating all possible values, namely
\begin{eqnarray}
e^{ - \frac{1}{K} \Y_{i,j} \left( \mathbf{b}_i(k)  \mathbf{b}_j(k) \right) } = c_{i,j} + c_{i,j}^\prime \cdot \Big( \mathbf{b}_i(k)  \mathbf{b}_j(k) \Big), \label{eqn:ee}
\end{eqnarray}
where $c_{i,j},c_{i,j}^\prime$ are two sample-specific constants, calculated by $c_{i,j} = \textstyle \frac{1}{2} ( e^{-\frac{1}{K} \Y_{i,j} } + e^{\frac{1}{K} \Y_{i,j}} )$ and $\textstyle c_{i,j}^\prime = \frac{1}{2} ( e^{-\frac{1}{K} \Y_{i,j}} - e^{\frac{1}{K} \Y_{i,j}} )$. Since the hardness of calculating the gradient of Eqn.~(\ref{eqn:ee}) lies in the bit product $\mathbf{b}_i(k)  \mathbf{b}_j(k)$, we replace the signum function using the sigmoid-shaped function $\sigma(\x) = 1/(1+\exp(-x))$, obtaining
\begin{eqnarray}
\mathbf{b}_i(k)  \mathbf{b}_j(k)   &=& \mbox{sign}(\w_k^\top \z_i) \cdot \mbox{sign}(\w_k^\top \z_j) \nonumber \\
&=& \mbox{sign}(\w_k^\top \z_i \z_j^\top \w_k) \nonumber \\
&\approx& 2 \cdot \sigma(\w_k^\top \z_i \z_j^\top \w_k) - 1. \label{eqn:relax}
\end{eqnarray}

Freezing the partial code product $\mathbf{b}_i(\setminus k) \circ \mathbf{b}_j(\setminus k)$, we define an approximate atomic loss with only bits $k$ active:
\begin{eqnarray}
\ell^{(k)}(\x_i,\x_j) &\triangleq& \textstyle e^{ - \Y_{i,j} \left( \mathbf{b}_i(\setminus k) \circ \mathbf{b}_j(\setminus k) \right) } \cdot \Big( c_{i,j} + \nonumber \\
&&  c_{i,j}^\prime \cdot ( 2 \cdot \sigma(\w_k^\top \z_i \z_j^\top \w_k) - 1 ) \Big),
\end{eqnarray}
where the first factor $e^{ - \Y_{i,j} ( \mathbf{b}_i(\setminus k) \circ \mathbf{b}_j(\setminus k) )}$ plays a role of re-weighting specific data pair, conditioned on the rest $K{-}1$ bits. Iterating over all $k$'s, the original loss function can now be approximated by
\begin{eqnarray}
\ell(\x_i,\x_j) \approx \textstyle \frac{1}{K} \sum_{k=1}^K \ell^{(k)}(\x_i,\x_j). \label{eqn:alossapprox}
\end{eqnarray}

\begin{algorithm}[tb]
   \caption{DeepHash Algorithm}
   \label{alg:deephash}
   \begin{small}
\begin{algorithmic}[1]
   \STATE {\bfseries Input:} Training set $\mathcal{X}$, data labels, and step size $\eta > 0$;
   \STATE {\bfseries Output:} network parameters $\w_k$, $k=1\cdots K$ for the hashing-loss layer, and $\boldsymbol\theta$ for other layers;
   \newline 
   {\bfseries \underline{pre-training stage \#1: initialize $\boldsymbol\theta$}}
   \vspace{0.04in}
   \STATE Concatenate all layers (excluding top hashing-loss layer) with a softmax layer that defines an image classification task;
   \STATE Apply AlexNet~\cite{KrizhevskySH12}) style supervised parameter learning algorithm, obtaining $\boldsymbol\theta$.
   \STATE Calculate neuron responses on second topmost layer through $\z = \phi(\x;\boldsymbol\theta)$;
   \newline 
   {\bfseries \underline{pre-training stage \#2: initialize $\w_k$}}
   \vspace{0.04in}
   \STATE Replicate all $\z$'s from previous stage;
   \WHILE{not converged}
   \STATE Forward computation starting from $\z$;
   \FOR{$k=1$ {\bfseries to} $K$}
    \STATE Update $\w_k$ by minimizing the image classification error;
    \ENDFOR
   \ENDWHILE
   \newline 
   {\bfseries \underline{simultaneous supervised fine-tuning}}
   \vspace{0.04in}
\WHILE{not converged}
\STATE Forward computation starting from the raw images;
\FOR{$k=1$ {\bfseries to} $K$}
    \STATE Estimate $\textstyle \partial \mathcal{Q} / \partial \w_k \propto \sum_{i,j,k} \partial \ell^{(k)}(\z_i,\z_j) / \partial \w_k$;
    \STATE Update $\w_k \leftarrow \w_k - \eta \cdot \partial \mathcal{Q} / \partial \w_k$;
    \ENDFOR
    \STATE Estimate $\textstyle \partial \mathcal{Q} / \partial \z_i \propto \sum_{j,k} \partial \ell^{(k)}(\z_i,\z_j) / \partial \z_i$, $\forall i$;
    \STATE Propagate $\textstyle \partial \mathcal{Q} / \partial \z_i$ to bottom layers, updating $\boldsymbol\theta$;
   \ENDWHILE
\end{algorithmic}
\end{small}
\end{algorithm}

Compared with other sigmoid-based approximations in previous hashing algorithms (\eg, KSH~\cite{liuw12}), ours only requires $|\w_k^\top \z_i \z_j^\top \w_k|$ (rather than both $|\w_k^\top \z_i|$ and $|\w_k^\top \z_j|$) is sufficiently large. This bilinearity-oriented relaxation is more favorable for reducing approximation error, which will be corroborated by the subsequent experiments.

Since the objective $\mathcal{Q}$ in Eqn.~(\ref{eqn:obj}) is a composition of atomic losses on data pairs, we only need to instantiate the gradient computation on specific data pair $(\x_i,\x_j)$. Applying basic calculus rules and discarding some scaling factors, we first obtain
\begin{eqnarray}
\frac{\partial \ell^{(k)}(\x_i,\x_j)}{\partial \w_k^\top \z_i \z_j^\top \w_k} &\propto& \textstyle e^{ - \Y_{i,j} \left( \mathbf{b}_i(\setminus k) \circ \mathbf{b}_j(\setminus k) \right) } \cdot c_{i,j}^\prime \nonumber \\
&& \textstyle \cdot \left( 1 -  \sigma(\w_k^\top \z_i \z_j^\top \w_k) \right) \cdot \sigma(\w_k^\top \z_i \z_j^\top \w_k), \nonumber
\end{eqnarray}
and further using calculus chain rule brings
\begin{eqnarray}
\frac{\partial \ell^{(k)}(\x_i,\x_j)}{\partial \w_k} &=& \frac{\partial \ell^{(k)}(\x_i,\x_j)}{\partial \w_k^\top \z_i \z_j^\top \w_k} \cdot \left( \z_i \z_j^\top + \z_j \z_i^\top \right) \w_k, \label{eqn:gw} \nonumber \\
\frac{\partial \ell^{(k)}(\x_i,\x_j)}{\partial \z_i} &=& \frac{\partial \ell^{(k)}(\x_i,\x_j)}{\partial \w_k^\top \z_i \z_j^\top \w_k} \cdot \left( \w_k \w_k^\top \z_j \right). \label{eqn:gz} \nonumber
\end{eqnarray}

Importantly, the formulas below obviously hold by the construction of $\ell^{(k)}(\x_i,\x_j)$:
\begin{eqnarray}
\frac{\partial \ell^{(k)}(\x_i,\x_j)}{\partial \w_{k'}} = \frac{\partial \ell^{(k)}(\x_i,\x_j)}{\partial \z_q} = \mathbf{0}, ~ k' \neq k, ~ q \neq i,j.
\end{eqnarray}

The gradient computations on other deep network layers simply follow the regular calculus rules. We thus omit the introduction.

\subsection{Two-Stage Supervised Pre-Training}
\label{subsection:pretrain}

Deep hashing algorithms (including ours) mostly strive to optimize pairwise (or even triplet as in~\cite{LaiPLY15}) similarity in Hamming space. This raises an intrinsic distinction compared with conventional applications of deep networks (such as image classification via AlexNet~\cite{KrizhevskySH12}). The total count of data pairs quadratically increases with regard to the training sample number, and in conventional applications the number of atomic losses in the objective only linearly grows. This entails a much larger mini-batch size in order to combat numerical instability caused by under-sampling\footnote{For instance, a training set with 100,000 samples demands a mini-batch of 1,000 data for $1\%$ sampling rate in image classification. In contrast, in deep hashing, capturing $1\%$ pairwise similarity requires a tremendous mini-batch of 10,000 data.}, which unfortunately often exceeds the maximal memory space on modern CPU/GPUs.

We adopt a simple two-stage supervised pre-training approach as an effective network pre-conditioner, initializing the parameter values in the appropriate range for further supervised fine-tuning. In the first stage, the network (excluding the hashing loss layer) is concatenated to a regular softmax layer. The network parameters are learned through optimizing the objective of a relevant semantics learning task (\eg, image classification). After stage one is complete, we extract the neuron outputs of all training samples from the second topmost layer (\ie, the variable $\z$'s in Section~\ref{subsec:loss}), feed them into another two-layer shallow network as shown in Figure~\ref{fig:network} and initialize the hashing parameters $\w_k$, $k=1\cdots K$. Finally, all layers are jointly optimized in a fine-tuning process, minimizing the hashing loss objective $\mathcal{Q}$. The entire procedure is illustrated in Figure~\ref{fig:network} and detailed in Algorithm~\ref{alg:deephash}.

\section{Experiments}

This section reports the quantitative evaluations between our proposed deep hashing algorithm and other competitors.

\begin{figure}[t]
\begin{center}
   \includegraphics[width=0.85\linewidth]{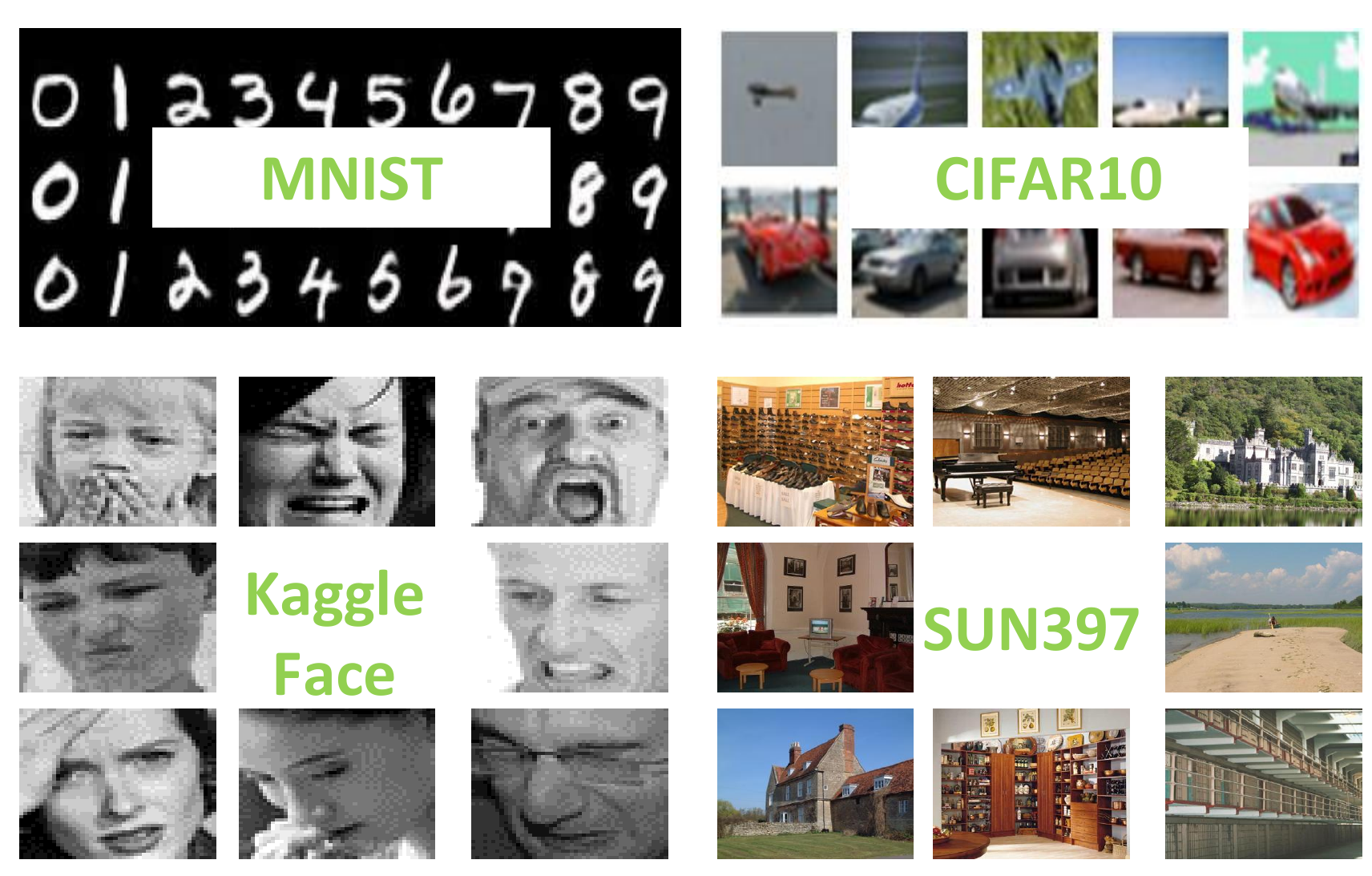}
\end{center}
   \caption{\small Exemplar images from MNIST, CIFAR10, Kaggle-Face and SUN397 datasets.}
\label{fig:example}
\vspace{-0.15in}
\end{figure}

\vspace{0.07in}
\noindent \textbf{Description of Datasets}: We conduct quantitative comparisons over four image benchmarks which represent different visual classification tasks. They include \textbf{MNIST}\footnote{\url{http://yann.lecun.com/exdb/mnist/}} for handwritten digits recognition, \textbf{CIFAR10}\footnote{\url{http://www.cs.toronto.edu/~kriz/cifar.html}} which is a subset of \emph{80 million Tiny Images} dataset\footnote{\url{http://groups.csail.mit.edu/vision/TinyImages/}} and consists of images from ten animal or object categories, \textbf{Kaggle-Face}\footnote{https://www.kaggle.com/c/challenges-in-representation-learning-facial-expression-recognition-challenge}, which is a Kaggle-hosted facial expression classification dataset to stimulate the research on facial feature representation learning, and \textbf{SUN397}~\cite{xiaoj10} which is a large scale scene image dataset of 397 categories. Figure~\ref{fig:example} shows exemplar images. For all selected datasets, different classes are completely mutually exclusive such that the similarity/dissimilarity sets as in Eqn~(\ref{eqn:y}) can be calculated purely based on label consensus. Table~\ref{table:data} summarizes the critical information of these experimental data, wherein the column of feature dimension refers to the neuron numbers on the second topmost layers (\ie, dimensions of feature vector $\z$).

\begin{table}[t]
\begin{center}
\begin{footnotesize}
\begin{tabular}{crrrr}
  \hline\hline
 \textbf{Dataset} & \textbf{Train/Query Set} &  \textbf{\#Class}   & \textbf{\#Dim} & \textbf{Feature}\\
  \hline
 \textbf{MNIST} & 50,000~/~10,000 &  10  & 500 & CNN\\
\hline
\textbf{CIFAR10} & 50,000~/~10,000 &  10  & 1,024 & CNN\\
\hline
\textbf{Kaggle-Face} & 315,799~/~7,178 &  7  & 2,304 & CNN\\
\hline
  \textbf{SUN397} & 87,003~/~21,751 &  397  & 9,216 & CNN\\
  \hline
\end{tabular}
\end{footnotesize}
\end{center}
\caption{\small Summary of the experimental benchmarks. Feature dimensions correspond to the neuron counts on the second topmost layer.}
\label{table:data}
\vspace{-0.1in}
\end{table}

\vspace{0.07in}
\noindent \textbf{Implementation and Model Specification}: We have implemented a substantially-customized version of the open-source Caffe~\cite{Jia13caffe}. The proposed hashing loss layer is patched to the original package and we also largely enrich Caffe's model specification grammar. Moreover, To ensure that mini-batches more faithfully represent the real distribution of pairwise affinities, we re-shuffle the training set at each iteration. This is approximately accomplished by using the trick of \emph{random skipping}, namely skipping the next few samples\footnote{The skipped samples vary at each operation, parameterized by a random integer uniformly drawn from $[0,200]$ in our experiments.} in the image database after adding one into the mini-batch.

\begin{table*}[t]
\begin{small}
\begin{center}
\begin{tabular}{|c||r|r|r|r|r|r|c|c|c|c|c|c|}
  \hline
  & \multicolumn{3}{|c|}{\textbf{MNIST}} & \multicolumn{3}{|c|}{\textbf{CIFAR10}} & \multicolumn{3}{|c|}{\textbf{Kaggle-Face}} & \multicolumn{3}{|c|}{\textbf{SUN397}} \\
 &\multicolumn{1}{|c}{12 bits} & \multicolumn{1}{c}{24 bits} & \multicolumn{1}{c|}{48 bits} &  \multicolumn{1}{c}{12 bits}  & \multicolumn{1}{|c}{24 bits} & \multicolumn{1}{c|}{48 bits} & \multicolumn{1}{c}{12 bits} &  \multicolumn{1}{c}{24 bits} & \multicolumn{1}{c|}{48 bits} & \multicolumn{1}{c}{12 bits} & \multicolumn{1}{c}{24 bits} &  \multicolumn{1}{c|}{48 bits}  \\
  \hline \hline
 \textbf{LSH}~\cite{charikar02} & 0.3717& 0.4933 & 0.5725&  0.1311 & 0.1619 & 0.2034 & 0.1911 & 0.2011 & 0.1976 & 0.0057 & 0.0060 & 0.0071\\
  \hline
  \textbf{ITQ}~\cite{GongLGP13} & 0.7578& 0.8132& 0.8293& 0.2711 &0.2825 & 0.2909 & 0.2435 & 0.2513 & 0.2514 &0.0268& 0.0361& 0.0471\\
  \hline
  \textbf{PCAH}~\cite{kulis09} & 0.4997& 0.4607& 0.3641& 0.2056 & 0.1867 & 0.1695 & 0.2169 & 0.2058 &0.1991 & 0.0218& 0.0261&  0.0315\\
  \hline
  \textbf{SH}~\cite{WeissTF08} &0.5175& 0.5330& 0.4898& 0.1935 & 0.1921 &0.1750 &0.2117 & 0.2054 & 0.2015 &0.0210& 0.0236&   0.0273\\
  \hline
  \textbf{LDAH}~\cite{Strecha12} & 0.5052 &0.3685 & 0.3093& 0.2187 & 0.1794 &0.1587 & 0.2154& 0.2032 &0.1961 & 0.0224& 0.0262&   0.0306\\
  \hline
\textbf{BRE}~\cite{BRE} &0.6950& 0.7498& 0.7785 & 0.2552 &0.2668 & 0.2864 & 0.2414 & 0.2522 & 0.2587 &0.0226 &0.0293 &   0.0372 \\
  \hline
\textbf{MLH}~\cite{NorouziF11} &0.6731 &0.4404& 0.4258& 0.1737 &0.1675 & 0.1737 & 0.2000 & 0.2115 & 0.2162 &0.0070 &0.0100 &   0.0210 \\
  \hline
\textbf{KSH}~\cite{liuw12} & 0.9537 &0.9713 & 0.9817& 0.3441 &0.4617 & 0.5482& 0.2862 & 0.3668 &0.4132 & 0.0194 & 0.0261&  0.0325\\
  \hline
\textbf{DH-1}~\cite{XiaPLLY14} & 0.957 &0.963& 0.960& 0.439 &0.511 & 0.522& -- & -- & -- & -- & -- &  --  \\
  \hline
\textbf{DH$\ast$-1}~\cite{XiaPLLY14} & 0.969 &0.975& 0.975& 0.465 &0.521& 0.532& -- & -- & -- & -- & -- &  --  \\
  \hline
\textbf{DH-2}~\cite{LiongLWMZ15} & 0.4675 & 0.5101 & 0.5250 & 0.1880 & 0.2083 & 0.2251 & -- & -- & -- & -- & -- &  --  \\
  \hline
\textbf{DH-3}~\cite{LaiPLY15} & -- & -- & -- & 0.552 & 0.566 & 0.581 & -- & -- & -- & -- & -- &  -- \\
  \hline
\textbf{DeepHash} & \textbf{0.9918} & \textbf{0.9931} & \textbf{0.9938} &\textbf{0.6874} & \textbf{0.7289}&  \textbf{0.7410} & \textbf{0.5487} & \textbf{0.5552} & \textbf{0.5615} & \textbf{0.0748}& \textbf{0.1054}&    \textbf{0.1293}\\
  \hline
\end{tabular}
\end{center}
\label{table:map1}
\caption{\small Experimental results in terms of mean-average-precision (mAP) under various hash bits. The mAP scores are calculated based on Hamming ranking. Best scores are highlighted in bold. Note that the mAP scores are in the numerical range of $[0,1]$. We directly cite the performance reported in~\cite{XiaPLLY14,LiongLWMZ15,LaiPLY15} since the source codes are not publicly shared. In the table, ``--" indicates the corresponding scores are not available. Refer to text for more details.}
\vspace{-0.1in}
\end{small}
\end{table*}

We designate the network layers for each dataset by referring to Caffe's model zoo~\cite{Jia13caffe}. Table~\ref{table:net} presents the deep network structure used for Kaggle-Face. The non-linear transform layers (\eg, RELU and local normalization layers) are ignored due to space limit. Specifically, the softmax layer is used only for pre-training the first 7 layers and not included during fine-tuning. We provide the network configuration information in the format of Caffe's grammar in the supplemental material.

\vspace{0.07in}
\noindent \textbf{Baselines and Evaluation Protocol}: All the evaluations are conducted on a large-scale private cluster, equipped with 12 NVIDIA Tesla K20 GPUs and 8 K40 GPUs. We denote the proposed algorithm as DeepHash. On the chosen benchmarks, DeepHash is compared against classic or state-of-the-art competing hashing schemes, including unsupervised methods such as random projection-based LSH~\cite{charikar02}, PCAH, SH~\cite{WeissTF08}, ITQ~\cite{GongLGP13}, and supervised methods like LDAH~\cite{Strecha12}, MLH~\cite{NorouziF11}, BRE~\cite{BRE}, and KSH~\cite{liuw12}. LSH and PCAH are evaluated using our own implementations. For the rest aforementioned baselines, we thank the authors for publicly sharing their code and adopt the parameters as suggested in the original software packages. Moreover, to make the comparisons comprehensive, four previous deep hashing algorithms are also contrasted, denoted as DH-1 and DH$\ast$-1 from~\cite{XiaPLLY14}, DH-2~\cite{LiongLWMZ15}, and DH-3~\cite{LaiPLY15}. Since the authors do not share the source code or model specifications, we instead cite their reported accuracies under identical (or similar) experimental settings.

Importantly, the performance of a hashing algorithm critically hinges on the semantic discriminatory power of its input features. Previous deep hashing works~\cite{XiaPLLY14,LaiPLY15} use traditional hand-crafted features (\eg, GIST and SIFT bag-of-words) for all baselines, which is not an optimal setting for fair comparison with deep hashing. To rule out the effect of less discriminative features, we strictly feed all baselines (except for four deep hashing algorithms from~\cite{XiaPLLY14,LiongLWMZ15,LaiPLY15}) with features extracted from some intermediate layer of the corresponding networks used in deep hashing. Specifically, after the first supervised pre-training stage in Algorithm~\ref{alg:deephash} is completed, we re-arrange the neuron responses on the layer right below the hashing loss layer (\eg, layer \#7 in Table~\ref{table:net}) into vector formats (namely the variable $\z$'s) and feed them into baselines.

All methods share identical training and query sets. After the hashing functions are learned on the training set, all methods produce binary hash codes for the querying data respectively. There exist multiple search strategies using hash codes for image search, such as hash table lookup~\cite{andoniI08} and sparse coding style criterion~\cite{LinSSHS14}. Following recent hashing works, we only carry out Hamming ranking once the hashing functions are learned, which refers to the process of ranking the retrieved samples based on their Hamming distances to the query. Under Hamming ranking protocol, we measure each algorithm using both mean-average-precision (mAP) scores and precision-recall curves.

\begin{figure*}[t]
  \centering
  \includegraphics[width=0.24\linewidth]{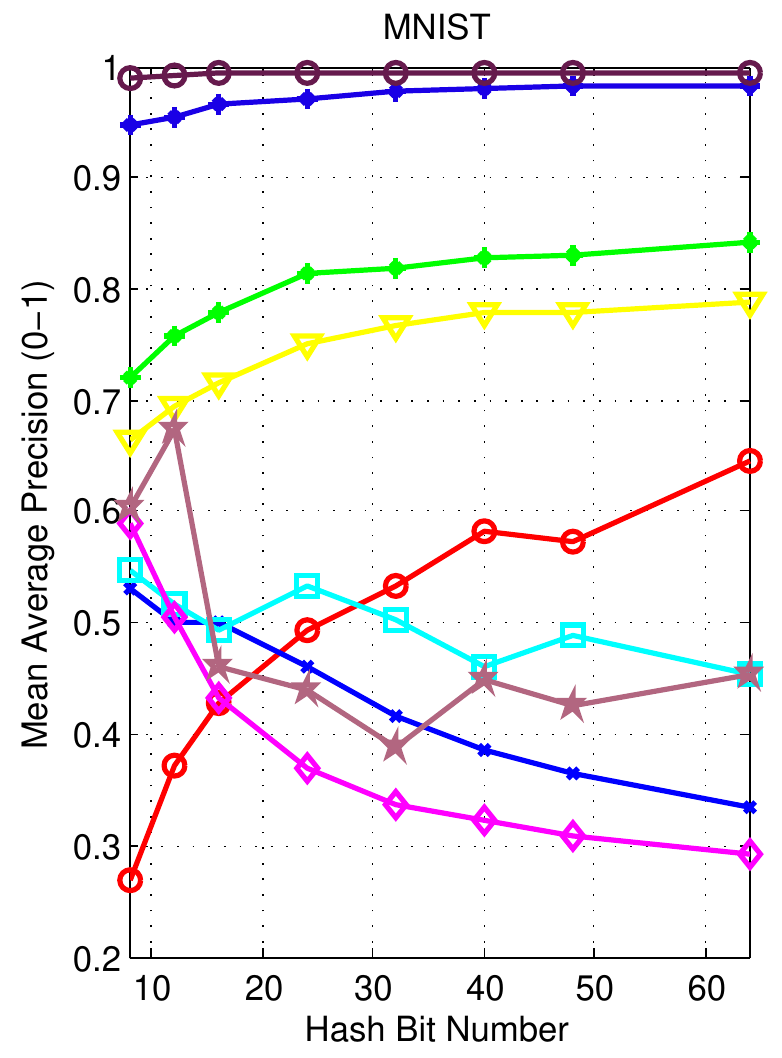}
\includegraphics[width=0.24\linewidth]{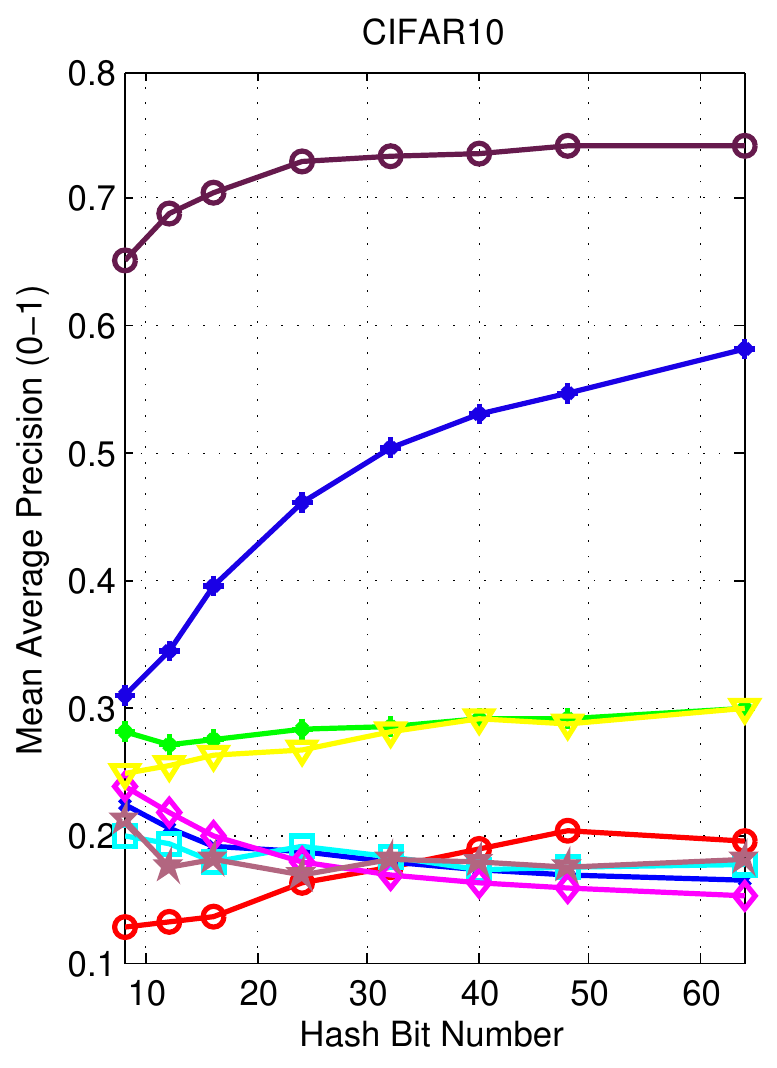}
  \includegraphics[width=0.24\linewidth]{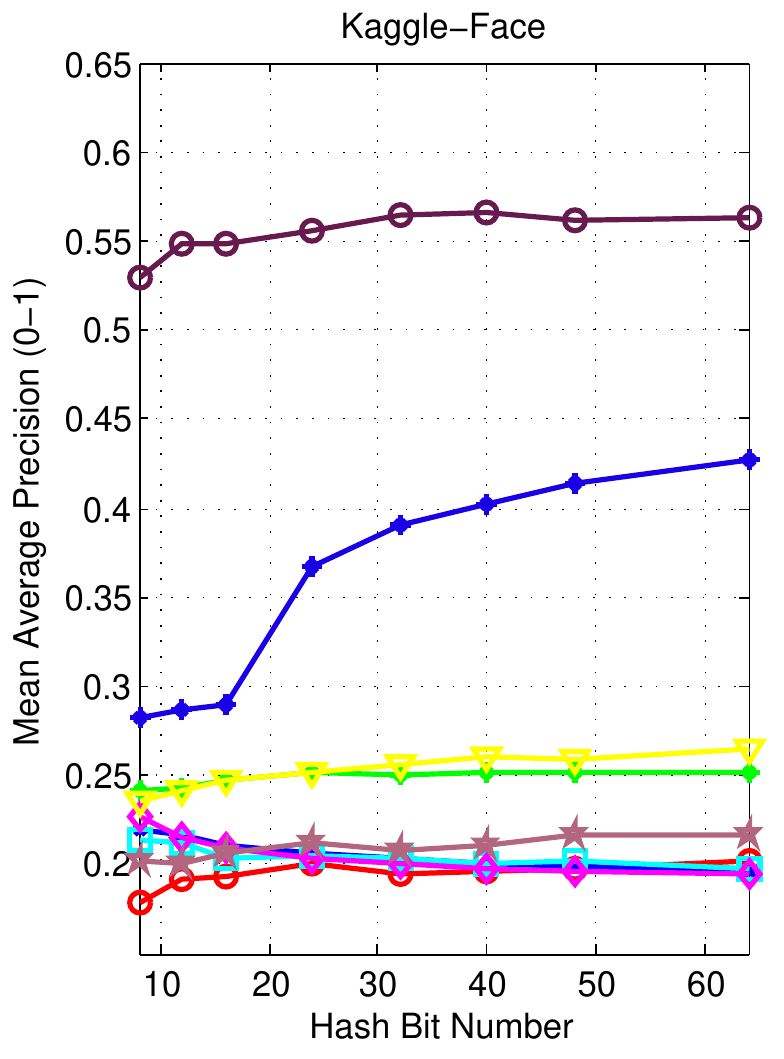}
  \includegraphics[width=0.24\linewidth]{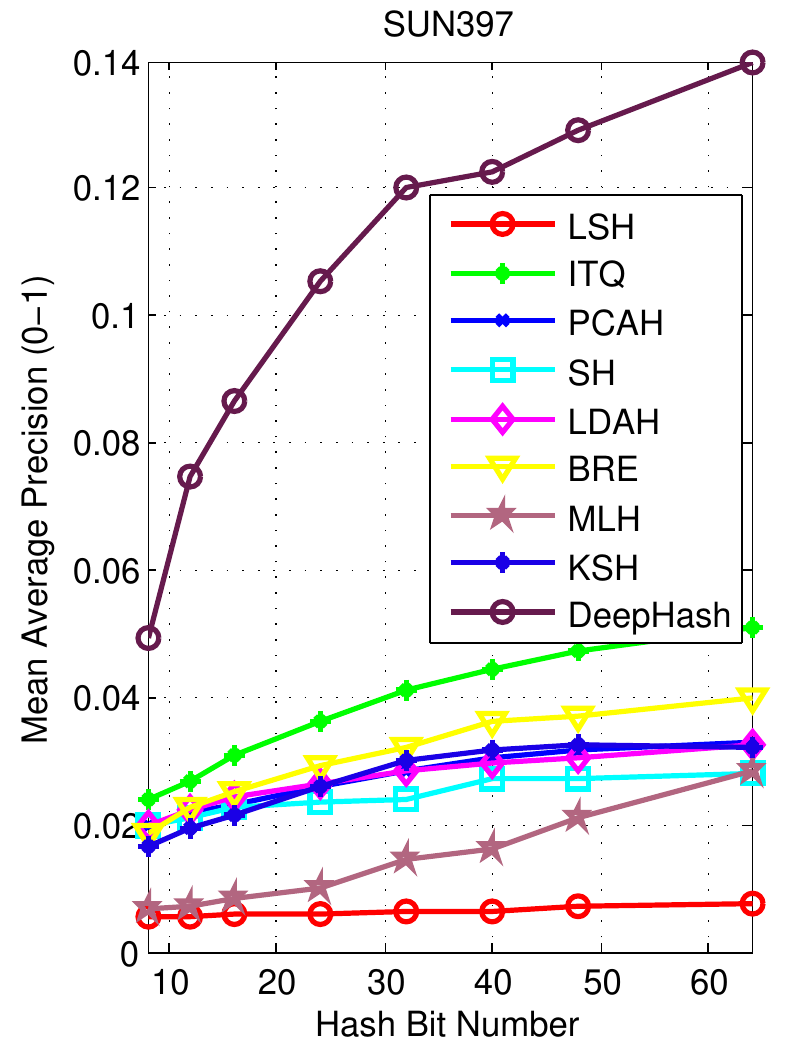}
  \caption{\small Experimental results in terms of mean-average-precision (mAP) under varying hash code lengths for all algorithms. Best viewing in color mode.}
  \label{fig:ap}
\end{figure*}

\begin{table*}[thb]
\begin{footnotesize}
\begin{center}
\begin{tabular}{|c||c|c|c|c|c|c|c|c|c|c|c|c|}
  \hline
  & \multicolumn{3}{|c|}{\textbf{MNIST}} & \multicolumn{3}{|c|}{\textbf{CIFAR10}} & \multicolumn{3}{|c|}{\textbf{Kaggle-Face}} & \multicolumn{3}{|c|}{\textbf{SUN397}} \\
 &\multicolumn{1}{|c}{12 bits} & \multicolumn{1}{c}{24 bits} & \multicolumn{1}{c|}{48 bits} &  \multicolumn{1}{c}{12 bits}  & \multicolumn{1}{c}{24 bits} & \multicolumn{1}{c|}{48 bits} & \multicolumn{1}{c}{12 bits} &  \multicolumn{1}{c}{24 bits} & \multicolumn{1}{c|}{48 bits} & \multicolumn{1}{c}{12 bits} & \multicolumn{1}{c}{24 bits} &  \multicolumn{1}{c|}{48 bits}  \\
  \hline \hline
 \textbf{DeepHash (random init.)} & 0.9806 & 0.9862 &  0.9873 &  0.5728 & 0.6503 & 0.6585 & 0.4125 & 0.4473 & 0.4620 & 0.0211 & 0.0384 & 0.0360\\
  \hline
  \textbf{DeepHash (pre-training)} & 0.9673 & 0.9753& 0.9796& 0.4986 & 0.5588 & 0.5966 & 0.4282 & 0.4484 & 0.4589 &0.0335& 0.0430& 0.0592\\
  \hline
  \textbf{DeepHash (fine-tuning)} & 0.9918 & 0.9931 & 0.9938 &0.6874 & 0.7289&  0.7410 & 0.5487 & 0.5552 & 0.5615 & 0.0748 & 0.1054&    0.1293 \\
  \hline
\end{tabular}
\end{center}
\label{table:finetune}
\caption{\small Comparisons of three strategies of parameter initialization and learning for the proposed DeepHash. See text for more details.}
\end{footnotesize}
\end{table*}

\begin{figure*}[t]
  \centering
  \includegraphics[width=0.24\linewidth]{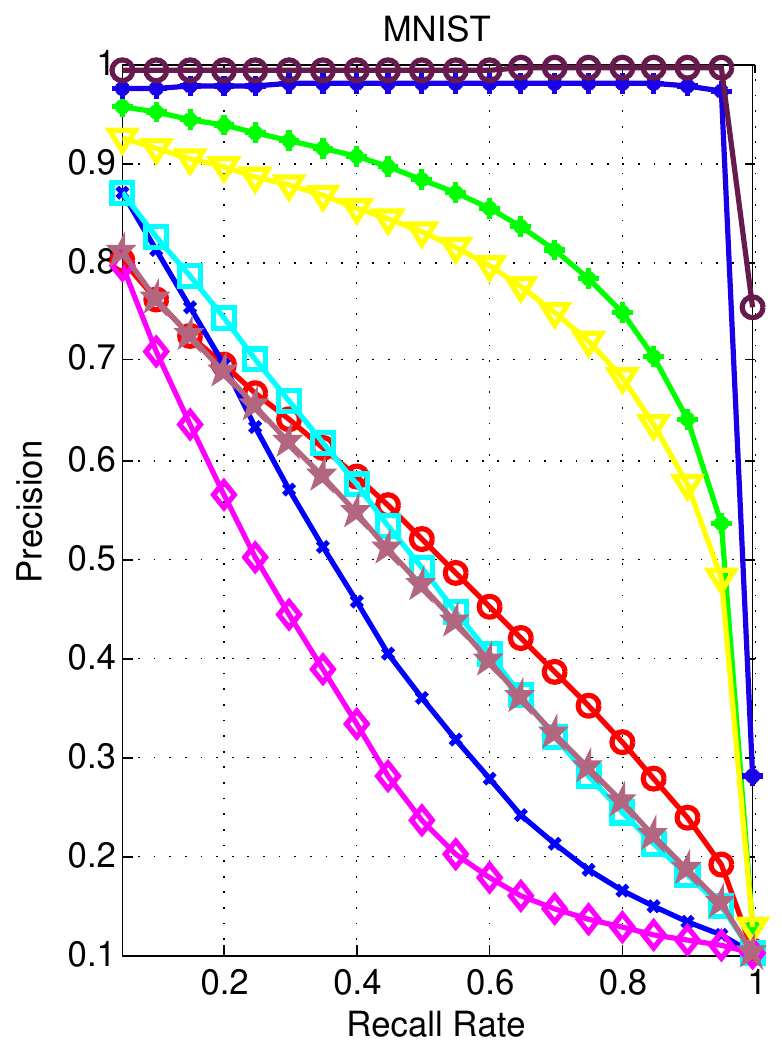}
\includegraphics[width=0.24\linewidth]{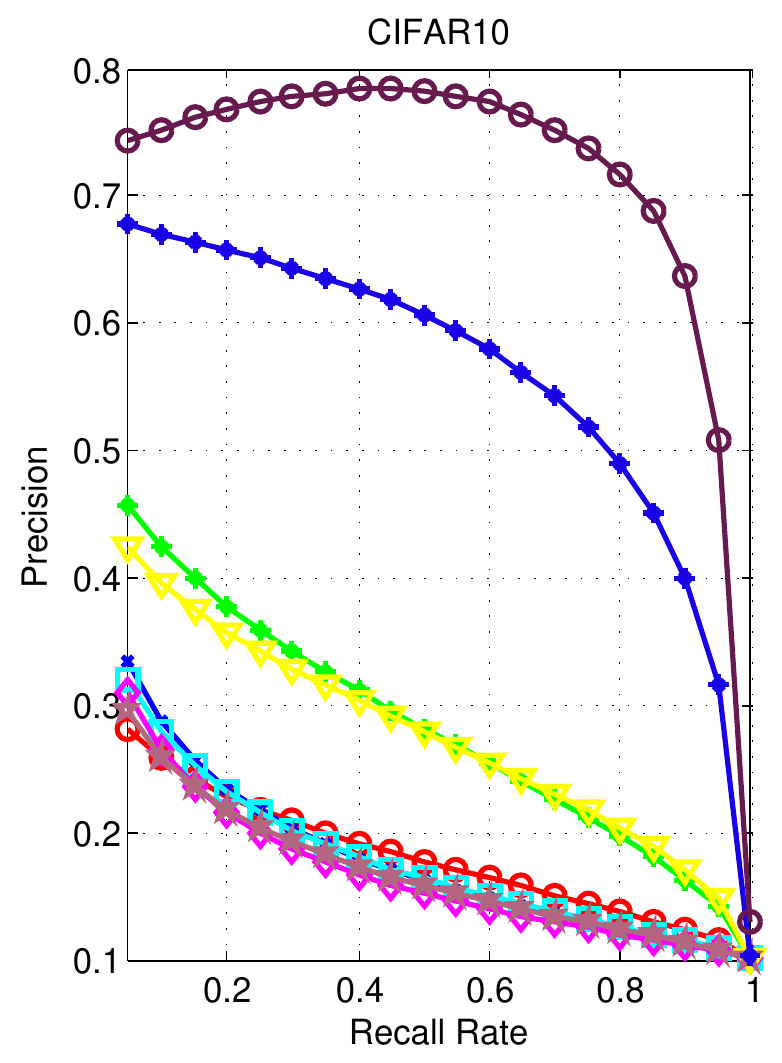}
  \includegraphics[width=0.24\linewidth]{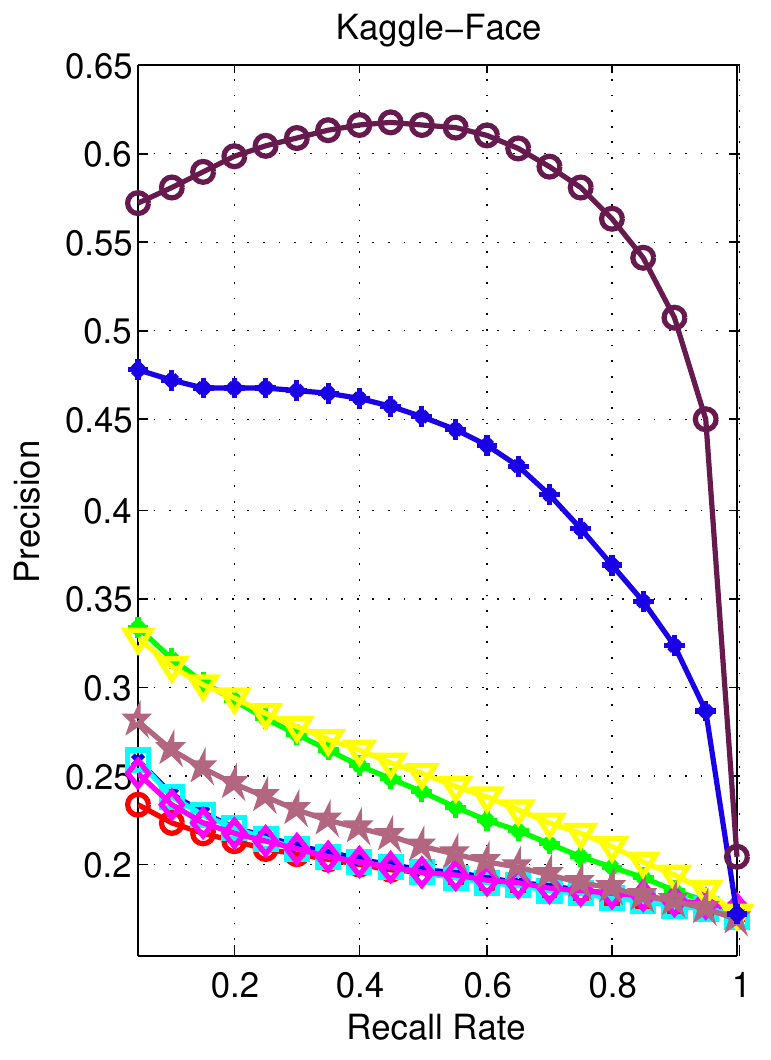}
  \includegraphics[width=0.24\linewidth]{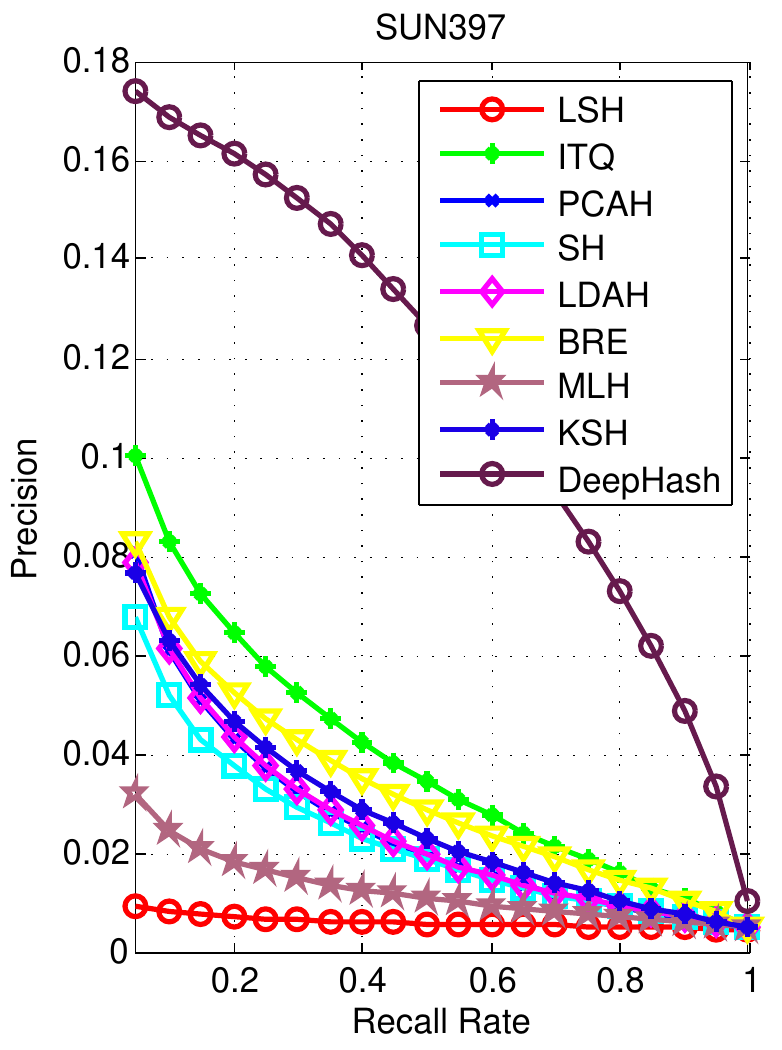}
  \caption{\small Precision-recall curves under 32 hash bits on all image benchmarks.}
  \label{fig:pr}
\end{figure*}

\vspace{0.07in}
\noindent \textbf{Investigation of Hamming Ranking Results}: Table~\ref{table:map1} and Figure~\ref{fig:ap} show the mAP scores for our proposed DeepHash algorithms (with supervised pre-training and fine-tuning) and all baselines. To clearly depict the evolving accuracies with respect to the search radius, Figure~\ref{fig:pr} displays the precision-recall curves for all algorithms with 32 hash bits. There are three key observations from these experimental results that we would highlight:

1) On all four datasets, our proposed DeepHash algorithm significantly perform better than all baselines in terms of mAP. For all non-deep-network based algorithm, KSH achieves the best accuracies on MNIST, CIFAR10 and Kaggle-Face, and ITQ shows top performances on SUN397. Using 48 hash bits, the best mAP scores obtained by KSH or ITQ are 0.9817, 0.5482, 0.4132, and 0.0471 on MNIST / CIFAR10 / Kaggle-Face / SUN397 respectively. In comparison, our proposed DeepHash performs nearly perfect on MNIST (0.9938), and defeat KSH and ITQ by very large margins, scoring 0.7410, 0.5615, and 0.1293 on other three datasets respectively.

2) We also include four deep hashing algorithms by referring to the accuracies reported in the original publications. Recall that the evaluations in~\cite{XiaPLLY14,LaiPLY15} feed baseline algorithms with non-CNN features (\eg, GIST). Interestingly, our experiments reveal that, when conventional hashing algorithms take CNN features as the input, the relative performance gain of prior deep hashing algorithms becomes marginal. For example, under 48 hash bits, KSH's mAP score 0.5482 is comparable with regard to DH-3's 0.581. We attribute the striking superiority of our proposed deep hashing algorithm to the importance of jointly conducting feature engineering and hash function learning (\ie, the fine-tuning process in Algorithm~\ref{alg:deephash}).

3) Elevating inter-bit mutual complementarity is overly crucial for the final performance. For those methods that generate hash bits independently (such as LSH) or by enforcing performance-irrelevant inter-bit constraints (such as LDAH), the mAP scores only show slight gains or even drop when increasing hash code length. Among all algorithms, two code-product oriented algorithm, KSH and our proposed DeepHash, show steady improvement by using more hash bits. Moreover, our results also validate some known insights exposed by previous works, such as the advantage of supervised hashing methods over the unsupervised alternatives.

\begin{table}[t]
\begin{center}
\begin{footnotesize}
\begin{tabular}{|c|c|c|c|c|}
  \hline
  & \multicolumn{2}{|c|}{\textbf{Hand-Crafted Feature}} & \multicolumn{2}{|c|}{\textbf{CNN Feature}}  \\
 &\multicolumn{1}{|c}{16 bits} & \multicolumn{1}{c|}{32 bits} & \multicolumn{1}{c}{16 bits} &  \multicolumn{1}{c|}{32 bits} \\
  \hline
 \textbf{LSH} & 0.1215 & 0.1385 &  0.1354 &  0.1752 \\
  \hline
  \textbf{ITQ} & 0.1528 & 0.1604 &  0.2757 &  0.2862 \\
  \hline
 \textbf{BRE} & 0.1308 & 0.1362 &  0.2634 &  0.2803 \\
 \hline
 \textbf{MLH} & 0.1373 & 0.1334 &  0.1810 &  0.1800 \\
 \hline
 \textbf{KSH} & 0.2191 & 0.2081 &  0.3958 &  0.5039 \\
 \hline
 \textbf{DeepHash} & 0.2166 & 0.2304 &  0.5472 &  0.5674 \\
  \hline
\end{tabular}
\end{footnotesize}
\end{center}

\begin{center}
\begin{footnotesize}
\begin{tabular}{|c|c|c|c|c|}
  \hline
  & \multicolumn{2}{|c|}{\textbf{Hand-Crafted Feature}} & \multicolumn{2}{|c|}{\textbf{CNN Feature}}  \\
 &\multicolumn{1}{|c}{16 bits} & \multicolumn{1}{c|}{32 bits} & \multicolumn{1}{c}{16 bits} &  \multicolumn{1}{c|}{32 bits} \\
  \hline
 \textbf{LSH} & 0.0067 & 0.0072 &  0.0059 &  0.0063 \\
  \hline
  \textbf{ITQ} & 0.0159 & 0.0157 &  0.0309 &  0.0410 \\
  \hline
 \textbf{BRE} & 0.0070 & 0.0075 &  0.0252 &  0.0319 \\
 \hline
 \textbf{MLH} & 0.0148 & 0.0147 &  0.0083 &  0.0144 \\
 \hline
 \textbf{KSH} & 0.0105 & 0.0095 &  0.0216 &  0.0300 \\
 \hline
 \textbf{DeepHash} & 0.0166 & 0.0189 &  0.0387 &  0.0525 \\
  \hline
\end{tabular}
\end{footnotesize}
\end{center}

\caption{\small mAP scores using hand-crafted features and CNN features in hashing-based image search. The method ``DeepHash" refers to the variant without fine-tuning. The top and bottom tables correspond to the results on CIFAR10 and SUN397 respectively.}
\label{table:nondeepfeature}
\end{table}

\vspace{0.07in}
\noindent \textbf{Effect of Supervised Pre-Training}: We now further highlight the effectiveness of the two-stage supervised pre-training process. To this end, in Table~\ref{table:finetune} we show the mAP scores achieved by three different strategies of learning the network parameters. The scheme ``DeepHash (random init.)" refers to initializing all parameters with random numbers without any pre-training. A typical supervised gradient back-propagation procedure as in AlexNet~\cite{KrizhevskySH12} is then used. The second scheme ``DeepHash (pre-training)" refers to initializing the network using two-stage pre-training in Algorithm~\ref{alg:deephash}, without any subsequent fine-tuning. It serves as an appropriate baseline for assessing the benefit of the fine-tuning process as in the third scheme ``DeepHash (fine-tuning)". In all cases, the learning rate in gradient descent drops at a constant factor (0.1 in all of our experiments) until the training converges.

There are two major observations from the results in Table~\ref{table:finetune}. First, simultaneous tuning all the layers (including the hashing loss layer) often significantly boosts the performance. As a key evidence, ``DeepHash (random init.)" demonstrates prominent superiority on MNIST and CIFAR10 compared with ``DeepHash (pre-training)". The joint parameter tuning of ``DeepHash (random init.)" is supposed to compensate the low-quality random parameter initialization. Secondly, positioning the initial solution near a ``good" local optimum is crucial for learning on challenging data. For example, the dataset of SUN397 has as many as 397 unique scene categories. However, due to the limitation of GPU memory, even a K40 GPU with 12GB memory only support a mini-batch of 600 samples at maximum. State differently, each mini-batch only comprises 1.5 samples per category on average, which results in a heavily biased sampling towards the pairwise affinities. We attribute the relatively low accuracies of ``DeepHash (random init.)" to this issue. In contrast, training deep networks with both supervised pre-training and fine-tuning (\ie, the third scheme in Table~\ref{table:finetune}) exhibit robust performances over all datasets.

\vspace{0.07in}
\noindent \textbf{Comparison with Hand-Crafted Features}: To complement a missing comparison in other deep hashing works~\cite{XiaPLLY14,LiongLWMZ15,LaiPLY15}), we also compare the hashing performance with conventional hand-crafted features and CNN features extracted from our second topmost layers. Following the choices in relevant literature, we extract 800-D GIST feature from CIFAR10 images, and 5000-D DenseSIFT Bag-of-Words feature from SUN397 images. The comparisons under 16 and 32 hash bits are found in Table~\ref{table:nondeepfeature}, exhibiting huge performance gaps between these two kinds of features. It clearly reveals how the feature quality impacts the final performance of a hashing algorithm, and a fair setting shall be established when comparing conventional and deep hashing algorithms.

\section{Concluding Remarks}

In this paper a novel image hashing technique is presented. We accredit the success of the proposed deep hashing to the following aspects: 1) it jointly does the feature engineering and hash function learning, rather than feeding hand-crafted visual features to hashing algorithms, 2) the proposed exponential loss function excellently fits the paradigm of mini-batch based training and the treatment in Eqn.~(\ref{eqn:alossapprox}) naturally encourages inter-bit complementarity, and 3) to combat the under-sampling issue in the training phase, we introduce the idea of two-stage supervised pre-training and validate its effectiveness by comparisons.

Our comprehensive quantitative evaluations consistently demonstrate the power of deep hashing for the data hashing task. The proposed algorithm enjoys both scalability to large training data and millisecond-level testing time for processing a new image. We thus believe that deep hashing is promising for efficiently analyzing visual big data.

\cleardoublepage
{
\bibliographystyle{ieee}
\bibliography{mybib}
}

\cleardoublepage

\section*{Appendix: Network Configurations}


Tables~\ref{table:net_mnist}-\ref{table:net_sun397} present the configurations of the deep networks used for selected benchmarks. The non-linear transform layers are majorly ReLU (rectified linear unit) and LRN (local response normalization). Specifically, the softmax layers are used only for pre-training the convolutional/innerProduct layers and not included during fine-tuning, which are thus not enumerated in these tables.

\begin{table}[h]
\begin{center}
\begin{small}
\begin{tabular}{crrr}
  \hline\hline
 \textbf{Layer ID} & \textbf{Layer Type} &  \textbf{Filter / Stride}   & \textbf{\#Dim of Output} \\
  \hline
 1 & data &  N/A  & $48\times48\times 3$ \\
\hline
2 & convolution &  $32\times 5\times5$ / 1 & $32\times48\times48$ \\
\hline
3 & max-pooling &  $3\times3$ / 2  & $32\times24\times24$ \\
\hline
4 & convolution &  $32\times 5\times5$ / 1 & $32\times24\times24$ \\
\hline
5 & avg-pooling &  $3\times3$ / 2  & $32\times12\times12$ \\
\hline
6 & convolution &  $64\times 5\times5$ / 1 & $64\times12\times12$ \\
\hline
7 & avg-pooling &  $3\times3$ / 2  & $64\times6\times6$ \\
\hline
 & soft-max &  N/A  & 10 \\
\hline
8 & hash-loss &  N/A  & \#(bit number) \\
\hline
\end{tabular}
\end{small}
\end{center}
\caption{\small Network configuration for the deep hashing task on Kaggle-Face. Information for other three benchmarks is found in the supplemental material.}
\label{table:net}
\end{table}

\begin{table}[h]
\begin{center}
\begin{small}
\begin{tabular}{crrr}
  \hline\hline
 \textbf{Layer ID} & \textbf{Layer Type} &  \textbf{Filter / Stride}   & \textbf{\#Dim of Output} \\
  \hline
 1 & data &  N/A  & $1\times 28\times28$ \\
\hline
2 & convolution &  $20\times 5\times5$ / 1 & $20\times24\times24$ \\
\hline
3 & max-pooling &  $2\times2$ / 2  & $20\times12\times12$ \\
\hline
4 & convolution &  $50\times 5\times5$ / 1 & $50\times8\times8$ \\
\hline
5 & max-pooling &  $2\times2$ / 2  & $50\times4\times4$ \\
\hline
6 & innerProduct & N/A & 500 \\
\hline
7 & ReLU & N/A & 500 \\
\hline
 & soft-max &  N/A  & 10 \\
\hline
8 & hash-loss &  N/A  & \#(bit number) \\
\hline
\end{tabular}
\end{small}
\end{center}
\caption{\small Network configuration for the deep hashing task on the MNIST digit data.}
\label{table:net_mnist}
\end{table}

\begin{table}[h]
\begin{center}
\begin{small}
\begin{tabular}{crrr}
  \hline\hline
 \textbf{Layer ID} & \textbf{Layer Type} &  \textbf{Filter / Stride}   & \textbf{\#Dim of Output} \\
  \hline
 1 & data &  N/A  & $3\times32\times32$ \\
\hline
2 & convolution &  $32\times 5\times5$ / 1 & $32\times32\times32$ \\
\hline
3 & max-pooling &  $3\times3$ / 2  & $32\times16\times16$ \\
\hline
4 & ReLU & N/A & $32\times16\times16$\\
\hline
5 & LRN & N/A & $32\times16\times16$ \\
\hline
6 & convolution &  $32\times 5\times5$ / 1 & $32\times16\times16$ \\
\hline
7 & ReLU & N/A & $32\times16\times16$ \\
\hline
8 & avg-pooling &  $3\times3$ / 2  & $32\times8\times8$ \\
\hline
9 & LRN & N/A & $32\times8\times8$ \\
\hline
10 & convolution &  $64\times 5\times5$ / 1 & $64\times8\times8$ \\
\hline
11 & ReLU & N/A & $64\times8\times8$ \\
\hline
12 & avg-pooling &  $3\times3$ / 2  & $64\times4\times4$ \\
\hline
 & soft-max &  N/A  & 7 or 10 \\
\hline
13 & hash-loss &  N/A  & \#(bit number) \\
\hline
\end{tabular}
\end{small}
\end{center}
\caption{\small Network configuration for the deep hashing tasks on Kaggle-Face and CIFAR10. The dimension of output in the softmax layer is 7 for Kaggle-Face and 10 for CIFAR10.}
\label{table:net_face}
\end{table}

\begin{table}[h]
\begin{center}
\begin{small}
\begin{tabular}{crrr}
  \hline\hline
 \textbf{Layer ID} & \textbf{Layer Type} &  \textbf{Filter / Stride}   & \textbf{\#Dim of Output} \\
  \hline
 1 & data &  N/A  & $3\times227\times 227$ \\
\hline
2 & convolution &  $96\times 11\times11$ / 4 & $96\times55\times55$ \\
\hline
3 & ReLU & N/A & $96\times55\times55$\\
\hline
4 & max-pooling &  $3\times3$ / 2  & $96\times27\times27$ \\
\hline
5 & LRN & N/A & $96\times27\times27$ \\
\hline
6 & convolution &  $256\times 5\times5$ / 1 & $256\times27\times27$ \\
\hline
7 & ReLU & N/A & $256\times27\times27$  \\
\hline
8 & max-pooling &  $3\times3$ / 2  & $256\times13\times13$ \\
\hline
9 & LRN & N/A & $256\times13\times13$ \\
\hline
10 & convolution &  $384\times 3\times3$ / 1 & $384\times13\times13$  \\
\hline
11 & ReLU & N/A & $384\times13\times13$ \\
\hline
12 & convolution &  $384\times 3\times3$ / 1 & $384\times13\times13$ \\
\hline
13 & ReLU & N/A & $384\times13\times13$ \\
\hline
14 & convolution &  $256\times 3\times3$ / 1 & $256\times13\times13$ \\
\hline
15 & ReLU & N/A & $256\times13\times13$ \\
\hline
16 & max-pooling &  $3\times3$ / 2   & $256\times6\times6$ \\
\hline
 & soft-max &  N/A  & 397 \\
\hline
17 & hash-loss &  N/A  & \#(bit number) \\
\hline
\end{tabular}
\end{small}
\end{center}
\caption{\small Network configuration for the deep hashing task on the SUN397 image benchmark.}
\label{table:net_sun397}
\end{table}

\end{document}